\documentclass{article}
\PassOptionsToPackage{numbers,sort&compress}{natbib}
\usepackage[preprint]{neurips_2026}
\usepackage[utf8]{inputenc}
\usepackage[T1]{fontenc}
\usepackage{courier}
\usepackage{hyperref}
\usepackage{url}
\usepackage{booktabs}
\usepackage{amsmath}
\usepackage{amsfonts}
\usepackage{graphicx}
\usepackage{microtype}
\usepackage{xcolor}
\usepackage{float}
\hypersetup{hidelinks,pdftitle={Finding the Right Fit: Model-Harness Interactions across Agent Tasks},pdfauthor={Yixuan Li, Yiyun Zhou, Yao Long Teng, Fuchao Yang, Yanchen Deng, Zhiyi Lyu, Xuyu Dong, Feng Chen, Bo An}}

\newif\ifdraftnotes
\draftnotesfalse

\newcommand{\draftgraphic}[3]{%
  \IfFileExists{#1}{\includegraphics[width=\linewidth]{#1}}{%
    \fcolorbox{gray!45}{gray!5}{\parbox[c][#2][c]{\dimexpr\linewidth-2\fboxsep-2\fboxrule\relax}{%
      \centering\small\textsf{#3}}}}}

\title{Finding the Right Fit:\\Model--Harness Interactions across Agent Tasks}
\author{%
  Yixuan Li \quad Yiyun Zhou \quad Yao Long Teng \quad Fuchao Yang \quad Yanchen Deng \\
  \textbf{Zhiyi Lyu \quad Xuyu Dong \quad Feng Chen \quad Bo An}\thanks{Corresponding author: \texttt{boan@ntu.edu.sg}.} \\
  College of Computing and Data Science, Nanyang Technological University, Singapore
}
\begin{document}
\maketitle
\begin{abstract}
Choosing an agent system means choosing both a language model and the harness through which it acts. We ask whether a strong model, harness, or pairing stays strong when the setting changes. We evaluate 66 configurations: four configurable harnesses (OpenHands, DeepSeek Harness, PI, and openJiuwen) paired with five models on TUA-Bench, ALE-CLI, and Terminal-Bench 4, plus the native \mbox{Codex--GPT} and \mbox{Claude Code--Claude} pairings. Model rankings reverse across harnesses. On Terminal-Bench 4, Claude leads GPT by 7.94 points in OpenHands but trails it by 30.16 points in PI. For four of the five models, the best harness changes from one benchmark to another, yet some pairings hold: openJiuwen gives Kimi its highest score on all three benchmarks, by 5.61 to 11.11 points. A model's own vendor harness is not reliably its best, and higher cost does not reliably buy a higher score. On Terminal-Bench 4, GPT scores higher under PI than under DSH at less than a quarter of the cost per task. Matched trajectories suggest why fit varies. Models start almost all repairs themselves, so much depends on whether the harness hands failures back in a form the model can use. GPT does best with PI's lean scaffold, while Kimi, which often issues malformed tool calls, does best in openJiuwen. We argue that the model, the harness, and the task should be evaluated together, and we release the harness adapters, evaluation code, and all 6,204 scored trajectories at \url{https://github.com/liyix/finding-the-right-fit} and \url{https://huggingface.co/datasets/yixuanli97/finding-the-right-fit}.

\end{abstract}

\section{Introduction}
A team deploying an agent faces two coupled choices: which model should reason, and which harness should turn its outputs into actions. A model leaderboard answers only the first. The harness decides how tools are presented, what stays in context, how errors come back to the model, and when the agent retries or stops. A model that does well in one harness can do much worse in another, and a harness that suits one model need not suit the next.

Prior work shows that harness choice affects task completion, efficiency, and failure behavior~\citep{yao2026harness,vats2026scaffold,kapoor2025hal}. The practical question is how far a good result carries over. Does a model ranking hold when the harness changes? Does a good pairing hold when the task changes? Does the model vendor's own harness settle the choice? A native harness is designed first for its vendor's product, so it need not be the best way to use the model on every workload, while third-party harnesses such as OpenHands aim to serve models from many vendors. These questions come up whenever a team picks a harness for an existing model, upgrades the model inside an existing agent, or moves a working agent to a new application.

We compare model--harness pairings on TUA-Bench, ALE-CLI, and a 63-task non-H100 subset of Terminal-Bench 4, which cover general terminal use, professional workflows, and hard command-line tasks. Four configurable harnesses, OpenHands, DeepSeek Harness (DSH), PI, and openJiuwen, are each paired with Claude Opus 5, GPT-6 Astra, GLM-5.3, Kimi K3, and DeepSeek V4 Pro, and the native Codex--GPT and Claude Code--Claude pairings serve as references, for 66 configurations in total. Every harness uses high reasoning effort and its own context management, and we report the three benchmarks separately.

The results show both shifting and persistent advantages. Claude beats GPT in OpenHands on Terminal-Bench but loses to it in the other three configurable harnesses, by as much as 30 points in PI. The best harness for Claude, GPT, GLM, and DeepSeek changes across benchmarks, whereas openJiuwen gives Kimi its best score on all three, and that lead comes from many tasks rather than a few outliers. Native pairings are mixed: Claude Code is Claude's best harness on TUA-Bench and ALE-CLI but trails OpenHands on Terminal-Bench, and Codex never gives GPT its highest score. Cost varies as much as score. Spending more does not reliably buy a higher score, and the same model can cost several times more under one harness than under another.

Matched trajectories help explain these differences. Models start almost all repairs themselves, so the deciding factor is often whether the harness returns a failure as feedback the model can use. A shell without a timeout turns a hung command into silence, and a stuck detector can end a run before a malformed call is fixed. A model that manages its own commands can do well with little scaffolding, as GPT does under PI, where it sets explicit timeouts on most of its shell calls. A model that makes more of these errors benefits from a harness that catches them, as Kimi does under openJiuwen. Many of the remaining failures happen at the very end, when the agent takes its own check as proof that the task is done.

Our central claim is that the value of a harness depends on the model and the task. We contribute a cross-benchmark matrix of 66 configurations with scores and costs, evidence that model rankings, best harnesses, and native advantages change with the setting while some pairings persist, trajectory analyses that link these differences to how harnesses handle feedback, continuation, and completion, and implications for training data and harness design. We also release adapters that run all six harnesses on the three benchmarks, the evaluation and cost-accounting code, and all 6,204 scored trajectories in a common format (Appendix~\ref{app:release}).

\section{Related Work}
\paragraph{Agent harnesses.}
Most language-model agents run a loop in which the model reasons, calls a tool, and reads the result~\citep{yao2023react}. The harness is everything around this loop: the tools and how they are exposed~\citep{schick2023toolformer,yang2024sweagent,wang2024codeact}, context and memory management~\citep{packer2023memgpt}, reusable skills~\citep{wang2023voyager,li2026mindskill}, planning and sub-agents, permission and sandbox rules, and the retry, timeout, and stopping policies that decide what happens when a step fails. OpenHands made many of these choices configurable in an open platform that serves models from many providers~\citep{wang2024openhands}, while Agentless reached competitive results on software repair with a fixed pipeline and no autonomous loop~\citep{xia2024agentless}. Model vendors now ship their own coding agents, such as Codex~\citep{codex} and Claude Code~\citep{claudecode}, alongside third-party harnesses such as OpenHands, PI~\citep{pi}, and DeepSeek Harness~\citep{dsh}, and agent frameworks such as openJiuwen~\citep{openjiuwen}. We compare existing harnesses under a common setup, without adding skills or memory beyond their defaults, rather than proposing a new one.

\paragraph{Benchmarks and cost-aware evaluation.}
Agent benchmarks now cover software repair~\citep{jimenez2024swebench}, interactive environments~\citep{liu2024agentbench}, computer use~\citep{xie2024osworld}, general terminal use~\citep{chen2026tua}, professional workflows~\citep{sun2026ale}, and hard command-line tasks~\citep{merrill2026terminal}. The cited Terminal-Bench paper describes version 2.0, while our campaign uses a non-H100 subset of version 4, and our local task subsets and mean-reward conventions are not direct reproductions of upstream leaderboard results. Many leaderboards fix either the scaffold or the model, which leaves the interaction between the two unmeasured. \citet{kapoor2024agents} argue that agent evaluations should report cost alongside accuracy, and HAL runs models and scaffolds across nine benchmarks, reports both, and uses log analysis to expose shortcuts and tool-calling failures~\citep{kapoor2025hal}. We follow this practice by pricing every configuration on a common basis.

\paragraph{Harness effects.}
Harness-Bench evaluates model--harness configurations on completion, process quality, efficiency, and failures, and finds that weaker models are more harness-sensitive and that output-contract violations are the largest failure class~\citep{yao2026harness}. The Scaffold Effect reports harness-specific failure patterns and up to 40-fold differences in tokens per solved task, with small pass-rate changes~\citep{vats2026scaffold}. The ALE-Claw analysis finds a larger spread across models than across harnesses and, in its tested settings, a lean harness that retains accuracy at lower cost~\citep{huang2026harness}. We add a full cross of four configurable harnesses with five models on three benchmarks, include vendor-native pairings, and ask which advantages survive a change of task.

\paragraph{Recovery, verification, and training.}
Language agents can improve by reflecting on feedback from earlier attempts~\citep{shinn2023reflexion}, but self-correction without external feedback is unreliable~\citep{huang2024selfcorrect}, and models often fail to verify their own outputs~\citep{hong2024closer,chen2026learning}. Our trajectories show the same split: repairs succeed when the harness returns a usable signal, while final self-checks against self-built criteria often miss real errors. On the training side, \citet{kim2026interplay} show that post-training with the harness in mind improves agents both in and out of distribution, while agents trained under a minimal harness degrade sharply when the tool environment shifts. Polar runs reinforcement learning through unmodified harnesses by proxying their model calls and reconstructing token-level trajectories, and its gains for the same model range from 0.6 points under the model's own Qwen Code harness to 22.6 points under Codex~\citep{xu2026polar}. We use matched trajectories to separate model-initiated, prompted, and system-performed steps before drawing implications for training data and harness design.

\section{Experimental Setup}
\subsection{Configurations and task collections}
The configurable grid contains $4\times5\times3=60$ reported entries. OpenHands~\citep{openhands}, DSH~\citep{dsh}, PI~\citep{pi}, and openJiuwen are each evaluated with the five models on all three task collections. Codex is evaluated only with GPT, and Claude Code only with Claude, adding six native references. Dashes in Table~\ref{tab:main} denote unevaluated pairings.

The intended task-set sizes are 120 for TUA-Bench (TUA), 99 for ALE-CLI (ALE), and 63 for Terminal-Bench 4 (TB4). ALE uses the campaign's local-Docker selection, and Terminal-Bench uses the shared non-H100 subset. The collections differ in workflow composition rather than forming orthogonal capability tests. All configurations use the same task identifiers.

Tasks affected by infrastructure failures were rerun. Each task counts once, using its last run for both reward and cost; a task without a valid outcome in that run receives zero, and partial rewards are retained. For openJiuwen--DeepSeek on ALE, image inputs in four image and computer-use tasks are served by DeepSeek V4.1 Flash, the provider's standard fallback for image requests (Appendix~\ref{app:audit}).

\subsection{Harness configuration}
\label{sec:harness_config}
We keep harness changes to a minimum, mainly what is needed to reach the models through a common endpoint. All models are served through OpenRouter and pinned to their first-party provider (Anthropic, OpenAI, Moonshot AI, DeepSeek, and Z.ai) with fallbacks disabled. Every harness requests high reasoning effort, although the same label need not imply the same thinking budget across harnesses. Sampling parameters are left at provider defaults, and context management, compaction, retries, and turn limits follow each harness's native defaults (Appendix Table~\ref{tab:harness_config}). Among the packaged harnesses, the one deliberate exception is OpenHands, whose context and output limits we set to 1M and 128K tokens because its defaults for a custom endpoint cap output at 16K. No configuration adds skills, MCP servers, memory, or custom prompts beyond its defaults. Harness tools are the native defaults; on ALE-CLI every harness also receives the benchmark's fourteen computer-use tools. Time limits are the benchmarks' own: 40 minutes per task for TUA-Bench, eight hours for Terminal-Bench, and the per-task limit of ALE-CLI (two hours for most tasks).

openJiuwen~\citep{openjiuwen} is an agent framework rather than a packaged coding agent and provides no official entry point for running these benchmarks. We therefore assembled a coding agent from its public harness API (v0.1.18), keeping the framework's native components wherever it provides them: seven coding tools (file read, write, and edit; glob; directory listing; grep; and a shell), its task loop, its context compression, and its security and model-anomaly rails. Skills, persistent memory, task planning, and sub-agents are disabled. On TUA-Bench and Terminal-Bench, a small added rail reports the remaining time limit before each model call. The same composition is used for every model and benchmark, without model- or task-specific prompts or tuning.

\subsection{Recorded reward and empirical fit}
For benchmark $b$ with intended task set $\mathcal T_b$ and size $N_b$, we report
\begin{equation}
 S_{h,m,b}=\frac{1}{N_b}\sum_{t\in\mathcal T_b}\widetilde r_{h,m,b,t},
 \qquad
 \widetilde r_t=\begin{cases}r_t,&\text{if a numeric verifier reward is retained},\\0,&\text{if the outcome is unresolved}.\end{cases}
 \label{eq:score}
\end{equation}
TUA and ALE allow partial credit, while Terminal-Bench is binary; we do not combine the three rewards into an overall score.

We use \emph{empirical fit} to describe which available harness has the highest recorded reward for a model and task collection:
\begin{equation}
 h^*(m,b)\in\operatorname*{arg\,max}_{h\in\mathcal H_{m,b}} S_{h,m,b},
 \label{eq:fit}
\end{equation}
where $\mathcal H_{m,b}$ contains only evaluated pairings. All score differences below are points on the 0--100 reward scale.

\subsection{Cost and trajectory analysis}
Cost is the agent-model API cost in USD at OpenRouter prices, including auxiliary model calls made by the harness and excluding evaluator calls. Costs, model-call counts, token counts, and agent time are taken from the same last run of each task that determines its reward. The trajectory analyses in Section~5 compare the final scored runs of the same task and model under different harnesses.

\section{Results}
\begin{figure}[p]
\centering
\includegraphics[width=\linewidth]{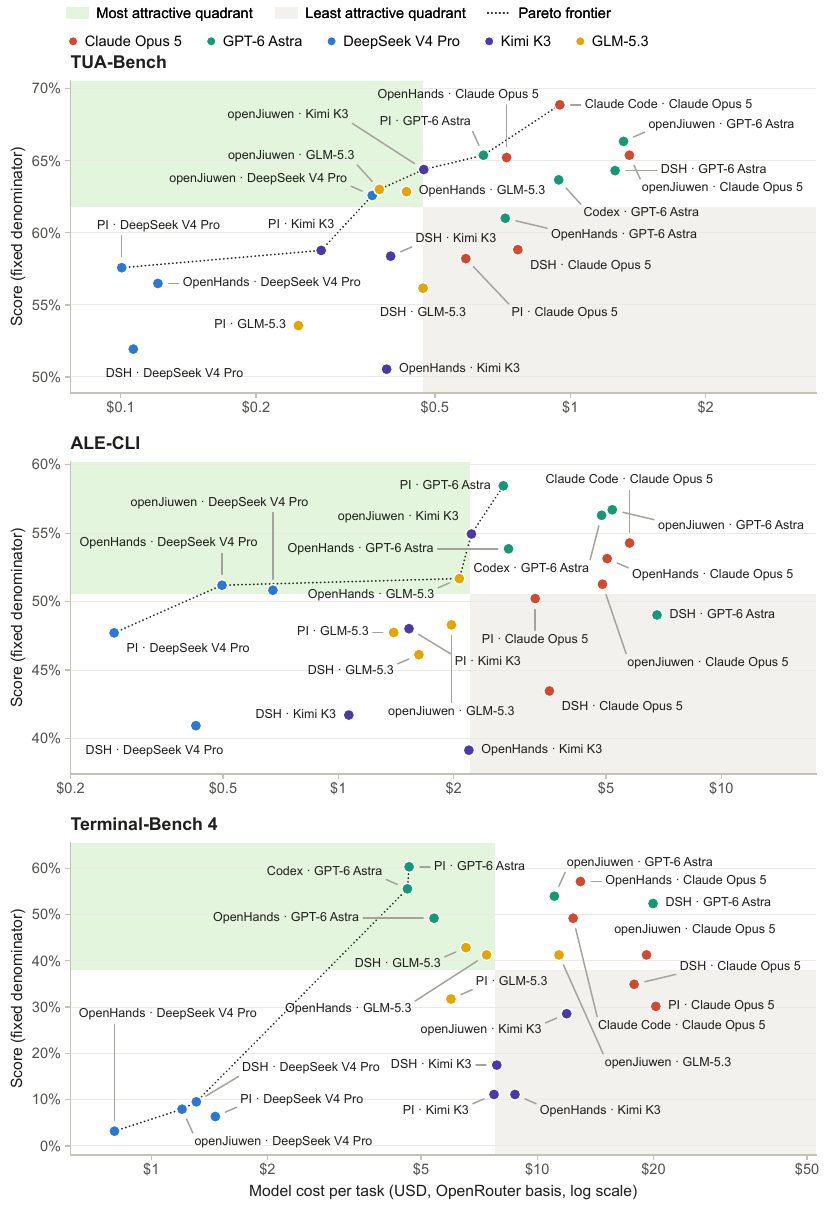}
\caption{Score versus model cost per task for all model--harness configurations on TUA-Bench, ALE-CLI, and Terminal-Bench~4. Cost per task uses OpenRouter prices (log scale). Shaded quadrants split each benchmark at the median cost and median score of its 22 configurations; the dotted line is the Pareto frontier.}
\label{fig:cost_performance}
\end{figure}

Table~\ref{tab:main} reports the rewards of all 66 configurations, and Figure~\ref{fig:cost_performance} places them against their cost per task.
\begin{table}[t]
\centering
\small
\caption{Mean rewards on fixed task-set denominators (120, 99, 63). Codex and Claude Code appear only with their implemented one-to-one model pairing. Bold marks the highest recorded score in each model--benchmark row; underline marks the second highest.}
\label{tab:main}
\setlength{\tabcolsep}{3.2pt}
\begin{tabular}{llrrrrrr}
\toprule
Benchmark & Model & OpenHands & DSH & PI & openJiuwen & Codex & Claude Code \\
\midrule
TUA-Bench & Claude Opus 5 & 0.6523 & 0.5884 & 0.5821 & \underline{0.6539} & --- & \textbf{0.6889} \\
 & GPT-6 Astra & 0.6101 & 0.6432 & \underline{0.6538} & \textbf{0.6635} & 0.6368 & --- \\
 & GLM-5.3 & \underline{0.6286} & 0.5616 & 0.5357 & \textbf{0.6301} & --- & --- \\
 & Kimi K3 & 0.5054 & 0.5838 & \underline{0.5878} & \textbf{0.6439} & --- & --- \\
 & DeepSeek V4 Pro & 0.5649 & 0.5193 & \underline{0.5758} & \textbf{0.6259} & --- & --- \\
\midrule
ALE-CLI & Claude Opus 5 & \underline{0.5314} & 0.4348 & 0.5022 & 0.5127 & --- & \textbf{0.5428} \\
 & GPT-6 Astra & 0.5386 & 0.4902 & \textbf{0.5846} & \underline{0.5671} & 0.5630 & --- \\
 & GLM-5.3 & \textbf{0.5168} & 0.4613 & 0.4775 & \underline{0.4830} & --- & --- \\
 & Kimi K3 & 0.3915 & 0.4172 & \underline{0.4802} & \textbf{0.5494} & --- & --- \\
 & DeepSeek V4 Pro & \textbf{0.5120} & 0.4094 & 0.4772 & \underline{0.5083} & --- & --- \\
\midrule
Terminal-Bench 4 & Claude Opus 5 & \textbf{0.5714} & 0.3492 & 0.3016 & 0.4127 & --- & \underline{0.4921} \\
 & GPT-6 Astra & 0.4921 & 0.5238 & \textbf{0.6032} & 0.5397 & \underline{0.5556} & --- \\
 & GLM-5.3 & \underline{0.4127} & \textbf{0.4286} & 0.3175 & \underline{0.4127} & --- & --- \\
 & Kimi K3 & 0.1111 & \underline{0.1746} & 0.1111 & \textbf{0.2857} & --- & --- \\
 & DeepSeek V4 Pro & 0.0317 & \textbf{0.0952} & 0.0635 & \underline{0.0794} & --- & --- \\
\bottomrule
\end{tabular}
\end{table}

\subsection{Model--Harness Interactions and Native Pairings}
\label{sec:interactions}

On Terminal-Bench 4, Claude records $36/63$ in OpenHands and GPT records $31/63$, a Claude advantage of 7.94 points. DSH reverses the ordering, with $22/63$ for Claude and $33/63$ for GPT. PI records $19/63$ and $38/63$, respectively, while openJiuwen records $26/63$ and $34/63$. The Claude-minus-GPT differences are $+7.94$, $-17.46$, $-30.16$, and $-12.70$ points. Thus the same models yield opposite rankings depending on the harness.

The reversal is especially clear when comparing OpenHands with PI on the same Terminal-Bench 4 task set. Claude drops from 57.14\% with OpenHands to 30.16\% with PI, whereas GPT rises from 49.21\% to 60.32\%, shifting the Claude-minus-GPT gap by 38.09 percentage points. This contrast shows that the observed model ordering is not simply preserved across harnesses.

Table~\ref{tab:native_comparison} compares each native pairing with the strongest evaluated alternative using the same model. With Claude, Claude Code leads the best alternative on TUA and ALE, by 3.50 and 1.14 points, whereas OpenHands leads Claude Code on Terminal-Bench by 7.94 points. For GPT, openJiuwen exceeds Codex by 2.67 points on TUA, while PI exceeds Codex by 2.15 points on ALE and 4.76 points on Terminal-Bench. Native provenance therefore does not consistently identify the best harness.

\begin{table}[t]
\centering\small
\caption{Native pairing versus the strongest observed alternative with the same model. $\Delta$ is alternative minus native reward on a 0--100 scale; negative values favor the native pairing. $\Delta$ is computed from unrounded scores.}
\label{tab:native_comparison}
\setlength{\tabcolsep}{4pt}
\begin{tabular}{llrlrr}
\toprule
Benchmark & Native pair & Score & Best alternative & Score & $\Delta$ \\
\midrule
TUA-Bench & CC--Claude & 0.6889 & openJiuwen & 0.6539 & -3.50 \\
TUA-Bench & Codex--GPT & 0.6368 & openJiuwen & 0.6635 & +2.67 \\
ALE-CLI & CC--Claude & 0.5428 & OpenHands & 0.5314 & -1.14 \\
ALE-CLI & Codex--GPT & 0.5630 & PI & 0.5846 & +2.15 \\
Terminal-Bench 4 & CC--Claude & 0.4921 & OpenHands & 0.5714 & +7.94 \\
Terminal-Bench 4 & Codex--GPT & 0.5556 & PI & 0.6032 & +4.76 \\
\bottomrule
\end{tabular}
\end{table}

\subsection{Task-Dependent Fit}
\label{sec:task_fit}

The highest-scoring harness changes across collections for four of five
models (Table~\ref{tab:fit}). GPT changes from openJiuwen on TUA-Bench to PI on
ALE-CLI and Terminal-Bench; GLM and DeepSeek change from openJiuwen to OpenHands
to DSH, while Claude changes from Claude Code on TUA-Bench and ALE-CLI to
OpenHands on Terminal-Bench. Kimi is the persistent pairing: openJiuwen scores
64.39\%, 54.94\%, and 28.57\% on the three collections, leading the strongest
observed alternatives by 5.61, 6.91, and 11.11 percentage points. The
changing winners and one stable favorable pairing both argue for evaluating fit
in the intended task setting.

\begin{figure}[!htbp]
\centering
\includegraphics[width=\linewidth]{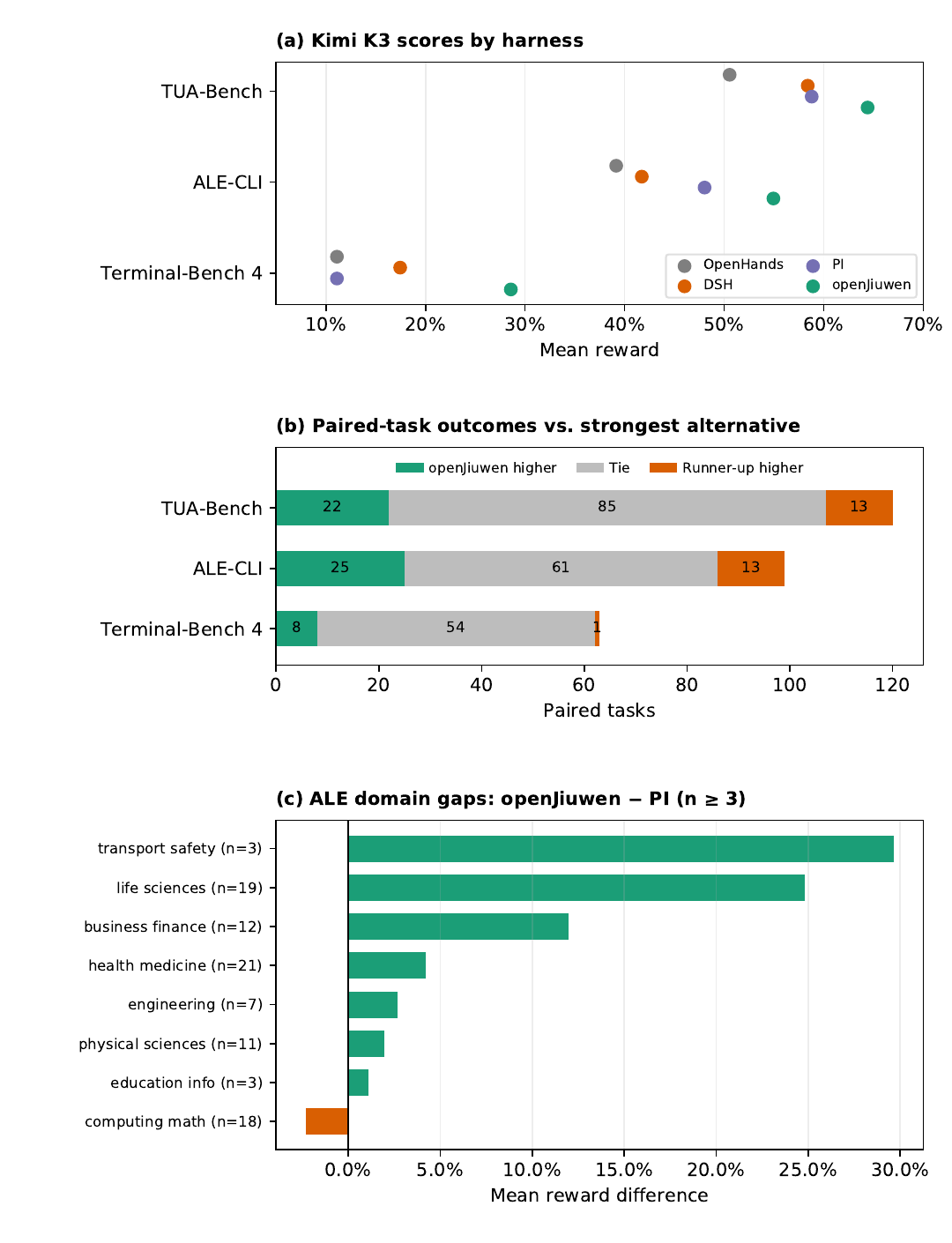}
\caption{Task-dependent fit for Kimi~K3. (a) Mean reward for the four
configurable harnesses on the fixed 120-, 99-, and 63-task sets. (b) Paired
task outcomes for openJiuwen versus the strongest observed alternative in each
collection (PI, PI, and DSH, respectively). (c) ALE-CLI mean paired differences
by domain, shown for domains with at least three tasks; positive values favor
openJiuwen.}
\label{fig:task_fit}
\end{figure}

The Kimi advantage is not attributable to a single outlier task. Against the
strongest alternative harness in each collection (PI on TUA-Bench and ALE-CLI,
and DSH on Terminal-Bench), openJiuwen is higher, tied, and lower on
22/85/13, 25/61/13, and 8/54/1 paired tasks, respectively
(Figure~\ref{fig:task_fit}). Removing the three largest positive task gaps still
leaves mean advantages of 3.11, 3.88, and 6.35 points. The ALE grouping further
localizes the effect: openJiuwen leads PI by 24.83 points over 19 life-science
tasks and by 11.99 points over 12 business/finance tasks, but trails by 2.32
points over 18 computing/math tasks. Thus the aggregate advantage is broad across tasks but not universal across
domains.

\begin{table}[t]
\centering\small
\caption{Highest-scoring available harness for each model and task collection. Native references are included. Four models change their observed winner across collections, while Kimi retains openJiuwen.}
\label{tab:fit}
\begin{tabular}{llll}
\toprule
Model & TUA-Bench & ALE-CLI & Terminal-Bench 4 \\
\midrule
Claude Opus 5 & Claude Code & Claude Code & OpenHands \\
GPT-6 Astra & openJiuwen & PI & PI \\
GLM-5.3 & openJiuwen & OpenHands & DSH \\
Kimi K3 & openJiuwen & openJiuwen & openJiuwen \\
DeepSeek V4 Pro & openJiuwen & OpenHands & DSH \\
\bottomrule
\end{tabular}
\end{table}

\subsection{Performance and Resource Use}
\label{sec:resources}
\paragraph{Performance and cost.}
Figure~\ref{fig:cost_performance} shows benchmark-dependent performance--cost trade-offs across model--harness configurations. TUA-Bench and ALE-CLI offer several intermediate trade-offs, whereas Terminal-Bench~4 shows a larger performance gap between the lowest recorded-cost configurations and the highest-scoring ones. Higher cost does not consistently yield better performance, and changing the harness can alter both score and cost for the same model. On Terminal-Bench~4, GPT-6 Astra scores 60.32\% with PI at \$4.66 per task but 52.38\% with DSH at \$19.94 per task, while PI--Claude costs \$20.28 per task for 30.16\%. These patterns support evaluating complete configurations in their target task setting.

\paragraph{Process efficiency.}
Same-model comparisons reveal different resource-use profiles across harnesses. Lower costs coincide with fewer model calls and fewer uncached input tokens, even when completion-token use increases (Appendix Figure~\ref{fig:prompt_tokens}). On ALE-CLI, Claude Opus~5 costs \$323 under PI and \$498 under OpenHands, with 3,888 versus 4,472 model calls and 119M versus 184M uncached input tokens, although it generates more completion tokens under PI (2.4M versus 2.1M). On Terminal-Bench~4, GPT-6 Astra makes 2,560 calls with 110M uncached input tokens under PI, compared with 11,879 calls and 673M under DSH. The harnesses also differ in how much context they reuse: openJiuwen serves 94--99\% of its input tokens from cache, compared with 49--77\% for the other harnesses (Appendix Figure~\ref{fig:resource_profile}). Trajectory length alone is insufficient to judge efficiency: solved tasks take more model calls than failed ones in 18 of 20 configurations on TUA-Bench and 17 of 20 on Terminal-Bench~4, but in only 1 of 21 on ALE-CLI, and agent time follows the same pattern (Appendix Figure~\ref{fig:turns_outcome}).

\section{Trajectory Analysis and Case Studies}
\subsection{Feedback and Recovery}
\label{sec:feedback}

We coded every actionable failure signal and the agent's response in ten same-task, same-model TB4 pairs
(OpenHands vs.\ PI, all five models; 192 events in final scored attempts), plus six Kimi pairs contrasting openJiuwen
with PI, and checked the resulting mechanisms against all final attempts on the three benchmarks.
Almost all responses were model-initiated (180 of 192). Diagnosis or a targeted patch was the most common
response (69\%), and it resolved the signal in 116 of these 133 cases; strategy changes were rarer but always resolved
the signal (16), and repetition was rare (11 events, 7 of them in
one run). Runs that passed answered feedback with diagnosis or repair more often than runs that failed (81\% vs.\ 56\%)
and almost never repeated themselves (1\% vs.\ 11\%). The decisive difference, however, was usually not whether the
model could repair an error but whether the harness returned the failure as usable feedback. OpenHands ends a run after
repeated identical failing actions: this ended 55 runs, 48 of them Kimi re-issuing an edit call without its required
content argument, and only one of these runs scored above zero (on TB4 \texttt{payments-\allowbreak pipeline-\allowbreak fix},
Kimi completed its design under OpenHands, then re-issued one malformed edit four times and was stopped with no code
written, while the same design passed under PI and openJiuwen). PI's shell has no default timeout, so 34 PI runs on TB4
and TUA sat blocked in a final command that never returned and produced no signal at all; the other harnesses bound
commands by default (see the matched case below). PI also ends a run when a turn is cut short by the output cap, whereas OpenHands continues and
openJiuwen re-prompts the model to continue with its partial reasoning kept (100 runs resumed, 48 of which went on to
score above zero). Of openJiuwen's harness prompts, only this resumption was decisive in the pairs we read (2 of 6);
its request to re-check all requirements before finishing triggered extra testing but did not decide any outcome.
When the decisive signal came from the agent's own test, failing runs in both harnesses tended to dismiss or skip it.
The fit is model-specific rather than a uniform ranking: GPT-6, which sets explicit timeouts on 56--67\% of its PI shell
calls, obtains its best TB4 and ALE scores under PI's minimal scaffold, while Kimi, penalized by the OpenHands editor and
by PI's unbounded and non-resuming behavior, scores highest under openJiuwen on all three benchmarks.
Turns cut off by provider stream errors (25 of Kimi's 62 PI runs on TB4) and sandbox stalls are scored as failures; we
attribute them to harness resilience rather than model behavior.

\begin{figure}[!htbp]
\centering
\includegraphics[width=\linewidth]{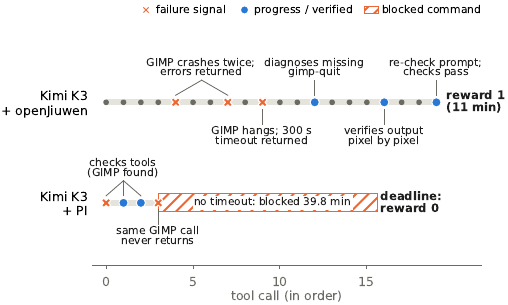}
\caption{Matched recovery case: Kimi~K3 on TUA \texttt{056-\allowbreak move-\allowbreak textbox-\allowbreak left}, final scored attempts. Both runs hit the
same GIMP hang. openJiuwen's shell returned it after 300\,s and the model diagnosed and recovered (reward~1); PI's shell
never returned, so the run received no feedback and ended at the deadline (reward~0). The timeout and the re-check prompt
come from the harness; all other actions are the model's.}
\label{fig:case_feedback}
\end{figure}

\paragraph{Matched recovery case.}
Kimi~K3 passed the TUA task \texttt{056-\allowbreak move-\allowbreak textbox-\allowbreak left} under openJiuwen (reward~1) and failed it
under PI (reward~0), although both runs hit the same bug (Figure~\ref{fig:case_feedback}). The model piped a GIMP script into
\mbox{\texttt{gimp -i -b -}} without a closing \texttt{(gimp-quit 0)}, which leaves GIMP waiting for input indefinitely.
Under PI, this was the model's first GIMP call after three setup commands. PI's shell has no default timeout and the model
set none, so the call never returned: the run received no further signal and ended at the 40-minute deadline with no
output file. Under openJiuwen, two earlier GIMP attempts crashed and the errors came back to the model, which moved its
script into a file and simplified it. A third attempt used the same construct and hung, but the shell's 300\,s limit
returned it as a timeout. The model then diagnosed the cause itself (``the script doesn't call gimp-quit''), found that
the output image had already been written, and verified it pixel by pixel (canvas size, background color, and the text
now on the left). When it declared completion, the harness's re-check prompt led to one more validation, which passed;
the run took 11 minutes. The harness supplied the signal and the model supplied the recovery: openJiuwen's bounded shell
turned the hang into feedback, while PI's unbounded shell turned the same hang into silence. Because the PI run never
received feedback, its failure is not evidence that Kimi cannot recover, and the openJiuwen run was partly fortunate that
the hung call had already written its output.

\subsection{Preserving Progress and Completing Deliverables}
\label{sec:delivery}

\begin{figure}[!tbp]
\centering
\vspace{-0.3cm}
\includegraphics[width=0.48\linewidth,trim=0 17 221 0,clip]{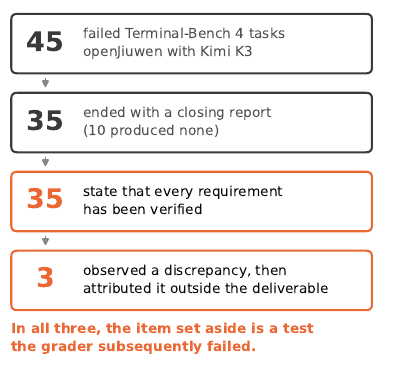}
\caption{Manual review of the 45 failed Terminal-Bench 4 tasks under openJiuwen with Kimi~K3. Every
closing report states that each specified requirement has been verified, and in three the agent had
observed a discrepancy and attributed it to a cause outside the deliverable.}
\label{fig:review}
\vspace{-0.3cm}
\end{figure}
We inspected all 45 failed Terminal-Bench 4 tasks for openJiuwen with Kimi K3 and found that the bottleneck lies at the end of the trajectory rather than in intermediate progress. Intermediate state is largely preserved: openJiuwen's context compression was triggered in only 1 of 63 Terminal-Bench 4 tasks, a rate we cross-checked on 120 TUA-Bench tasks and again observed only once. In the sole Terminal-Bench 4 case, \texttt{layout-config-recreation}, compression reduced 1,450 messages to three ($\sim$834k to $\sim$7.8k tokens) while preserving the deliverable path, measured scores, rejected alternatives, and an explicit resume point.

\begin{figure}[!tbp]
\centering
\includegraphics[width=\linewidth]{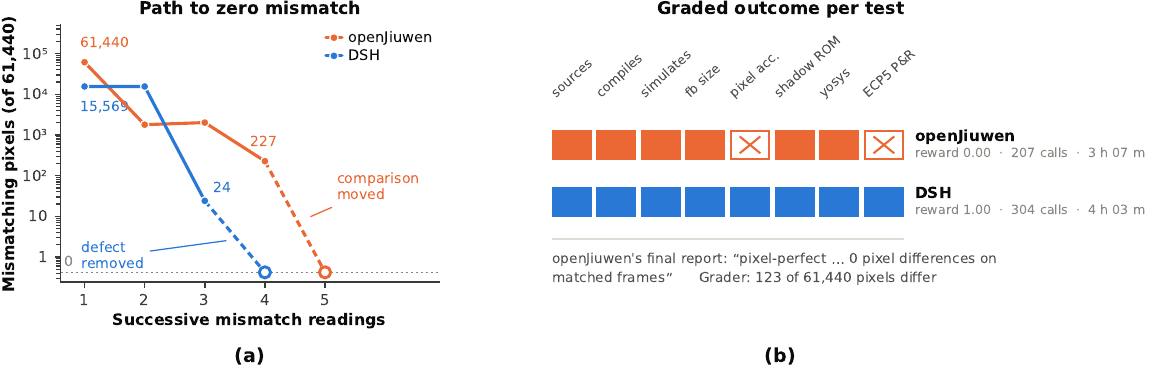}
\vspace{-0.7cm}
\caption{Matched progress-to-delivery case on \texttt{retro-console-soc} with Kimi~K3, where openJiuwen scores 0.00 and DSH scores 1.00. \textbf{(a)} Successive mismatch readings that each configuration obtained against its own acceptance criterion. Readings are consecutive within a run and are not aligned in time across runs; open markers are exact zeros, drawn on the axis floor, and the dashed segment marks the step that reaches zero. openJiuwen reaches zero by comparing its own console model at frame~60 against the RTL output at frame~40, whereas DSH reaches it by removing the remaining defect. \textbf{(b)} Per-test verifier outcome (filled, pass; crossed, fail) with the reward, tool calls, and wall-clock time of each run. The property asserted by openJiuwen's final report is the one the grader rejects.}
\label{fig:case_delivery}
\end{figure}

What is missing instead is a reliable mechanism for verifying completion. Of the 45 failures, 35 ended
with a final report, and our review of each one found that all 35 state that every specified
requirement has been verified (Figure~\ref{fig:review}). Three did not lack a signal: the agent
observed a discrepancy while self-checking but attributed it to a cause outside the deliverable, namely
a defect in its own test harness, a fix already applied in the current run, or pre-existing upstream
behavior, and stopped treating it as unresolved. This is a characteristic self-verification
failure~\cite{hong2024closer,chen2026learning}; in all three, the item dismissed is a test the grader
subsequently failed.

The failure is structural. No harness exposes the acceptance criterion to the agent during the run,
so every configuration must construct its own, and the evidence a final check rests on is therefore
produced by the agent itself. The 35 self-certified reports show that checking cannot substitute for
supplying the criterion against which to check. Nor is the pattern unique to openJiuwen: on \texttt{layout-config-recreation}, PI with the same model failed the same two tests.

\paragraph{Matched progress-to-delivery case.}

Terminal-Bench 4's \texttt{retro-console-soc} requires a synthesizable 8-bit game-console SoC whose
simulated video output must match the reference frame pixel by pixel. The task ships only a test ROM,
so both configurations had to construct their own acceptance criterion; what separates them is how they
used it. On Kimi~K3, openJiuwen scores 0.00 and DSH scores 1.00 (Figure~\ref{fig:case_delivery}).

Both drove their measured mismatch to zero, openJiuwen after genuine repairs that cut it by more than
two orders of magnitude (Figure~\ref{fig:case_delivery}(a)). It did not close the remaining gap,
however, but moved the comparison, holding the RTL output at one frame while advancing its own model to
another, and read the resulting zero as proof of pixel-exactness. A final confirmation step
then reused the same criterion and returned no new issues; the final report asserted
pixel-perfect output. Comparing the submitted framebuffer against the withheld reference, the grader
rejected precisely that property (Figure~\ref{fig:case_delivery}(b)).

DSH reached zero by removing the defect itself and passed every test, on a longer trajectory than
openJiuwen's. Trajectory length therefore does not separate the two; what separates them is whether the
final zero was obtained by repairing the deliverable or by changing what was compared.

\section{Implications for Data, Training, and Harness Design}
\paragraph{What a trajectory teaches.}
A scored trajectory is not a uniform record of model behavior. Each consequential step has an initiator: the model acting on its own, the model acting after a harness message, or the harness acting on the model's behalf. This label changes what the step can teach. Model-initiated repair is the clearest training signal. Almost all responses to failure signals in the Terminal-Bench pairs were model-initiated (Section~\ref{sec:feedback}), and on \texttt{payments-\allowbreak pipeline-\allowbreak fix} the same Kimi~K3 run under PI diagnosed and fixed four self-detected failures. Harness-performed operations are different. When OpenHands' stuck detector ends a run after repeated malformed edit calls, the model receives no final message, and when a PI command never returns, the trace ends without any signal. Such endings record the outcome of the configuration, but they are weak evidence about what the model would have done with feedback. Prompted decisions need the most care. On TUA task \texttt{042-\allowbreak black-\allowbreak sale-\allowbreak coffee-\allowbreak makers}, GPT-6 Astra hit a Google reCAPTCHA under openJiuwen, PI and DSH and, in all three, stopped and asked the user to complete it. PI and DSH ended there with reward~0. openJiuwen, whose completion rail returns a generic ``Continue working on the original task'' message when a round ends without a completion marker, sent that message twice. The model then installed an offline speech recognizer, transcribed the audio challenge, passed the check and completed the task (reward~1). The continuation message fired on only 9 of 120 GPT TUA tasks, and this was the only one of the 9 where it changed the outcome relative to both PI and DSH. The case shows that outcome reward alone is a poor filter for imitation data. The highest-reward trace here contains a prompted decision to defeat an anti-automation check, while the lower-reward traces contain the hand-back a deployment policy may prefer. The delivery cases show a related gap. On \texttt{retro-\allowbreak console-\allowbreak soc} and \texttt{layout-\allowbreak config-\allowbreak recreation}, openJiuwen--Kimi replaced the stated acceptance criterion with a check of its own and reported success against it (Section~\ref{sec:delivery}). The decision worth recording is therefore the choice of acceptance test, together with the evidence cited for completion, and not only the final artifact. These are proposals for how to annotate trajectories. We did not train on them.

\paragraph{What to train and what to support at runtime.}
The traces separate behaviors that a generic harness message did not supply from failures that a harness default produced. Choosing an acceptance test that matches the task, and noticing when a self-built oracle has drifted from it, belong in the first group. openJiuwen asks the model to re-check every stated requirement before it finishes. This prompt fired on 54 of 63 Kimi Terminal-Bench tasks and on almost every TUA task, yet it did not decide the outcome in any matched pair we read. On \texttt{retro-\allowbreak console-\allowbreak soc}, the model spent seven more turns re-checking with the same self-built oracle and again reported pixel-exact output. The grader found 123 of 61440 pixels wrong. A reminder that does not change the evidence a model consults is unlikely to substitute for this judgment, which makes it a candidate for training on contrastive pairs such as the successful DSH run on the same task. Tool-argument adherence is a second candidate. Kimi's repeated edit calls without required arguments ended most OpenHands stuck-detector runs, but the same model avoided them under openJiuwen's separate write and edit tools. The agent RL framework Polar reports post-training gains that differ widely across harnesses for the same model, from 0.6 to 22.6 points~\citep{xu2026polar}, but we did not test training for our models. Other failures come from harness defaults and are cheaper to fix at runtime. These include bounded command timeouts that return a signal, argument errors that are returned to the model rather than silently ending the run, and continuation after a turn that stops early. The CAPTCHA case adds one requirement for continuation. A generic continue message cannot tell an unfinished step from one that should be escalated to a human. Continuation therefore needs a way for the model to report that a step requires a person, and a harness policy that respects that report. The broader implication is that no single model--harness pairing is the default choice. The highest-scoring harness changes across collections for four of five models (Table~\ref{tab:main}). Native provenance does not settle the choice either (Section~\ref{sec:interactions}). Claude Code leads the Claude comparisons on TUA and ALE, but OpenHands leads on Terminal-Bench by 7.94 points. Codex never records the highest GPT score; openJiuwen or PI exceeds it on each collection. A more natural practice is to select the model and harness for the benchmark and task at hand, and to evaluate each runtime support within that configuration. A configurable platform can support this practice. openJiuwen records the highest TUA score for four of five models and the highest Kimi score on all three collections, and its continuation, completion and budget behaviors are separate components that can be adjusted per configuration. Which support matters depends on a model's habits rather than its overall strength: the most consequential continuation event involved GPT, while the tool-argument loop involved Kimi.

\begin{table}[t]
\centering\small
\caption{Evidence-grounded implications. Initiator: M = model-initiated, P = after a harness prompt, S = performed by the system.}
\label{tab:implications}
\begin{tabular}{p{0.30\linewidth}p{0.30\linewidth}p{0.30\linewidth}}
\toprule
Observed behavior (initiator) & Data / training implication & Runtime harness implication \\
\midrule
Kimi re-issues edit calls without required arguments until OpenHands' stuck detector ends the run; the same model repairs failures under PI (TB4; M, then S) &
Collect malformed call $\rightarrow$ error $\rightarrow$ corrected call; train argument adherence. Do not treat the system termination as a model failure to recover &
Return argument errors as structured feedback and warn before terminating; offer separate write and edit tools \\
\addlinespace
All three GPT runs hand a CAPTCHA to the user; only openJiuwen's generic continue message leads to solving the audio challenge (TUA 042; M, then P) &
Label prompted decisions; do not select imitation data by reward alone; keep policy-compliant hand-backs as positive examples &
Scope continuation prompts; provide an explicit escalation channel for steps that need a human \\
\addlinespace
Kimi substitutes its own acceptance check and reports success; the re-check round reuses the same oracle (TB4 \texttt{retro-console-soc}; M, then P) &
Contrastive pairs (DSH success, same task) at the acceptance-test decision; train alignment of evidence with stated criteria &
A generic re-check prompt is insufficient; expose checkable deliverable criteria where the task permits \\
\bottomrule
\end{tabular}
\end{table}

\section{Limitations}
The study compares model--harness configurations on three fixed task subsets with one counted run per task. Model settings, tools, and execution budgets are not uniformly matched, so the observed differences do not isolate the causal effect of any single component, and run-to-run variance is not measured. The trajectory analyses rest on a small number of matched pairs, and the data and harness implications we draw from them are proposals rather than tested improvements.

\section{Conclusion}
Across 66 configurations, model rankings, the best harness for each model, and the standing of native pairings all depend on the task setting, while some pairings, such as openJiuwen with Kimi, stay ahead on all three benchmarks. Cost changes with the harness as much as score does, and matched trajectories tie much of this variation to how a harness returns failures, continues interrupted work, and decides when a task is finished. Model--harness fit is therefore a practical unit for agent evaluation and deployment. Identifying which harness components drive a given fit, and choosing configurations automatically for unseen tasks, are natural next steps.

\clearpage
\bibliographystyle{plainnat}
\bibliography{references}
\clearpage
\appendix
\small
\section{Run details}
\label{app:audit}

\paragraph{Harnesses and models.}
Table~\ref{tab:harness_config} lists the harness versions and their native tools, context management, and turn limits. TUA-Bench and Terminal-Bench run through Harbor v0.22.0 and ALE-CLI through its own runner. The OpenRouter model identifiers and pinned providers are \texttt{anthropic/claude-opus-5} (Anthropic), \texttt{openai/gpt-6-astra} (OpenAI), \texttt{moonshotai/kimi-k3} (Moonshot AI, FP4), \texttt{deepseek/deepseek-v4-pro-0813} (DeepSeek), and \texttt{z-ai/glm-5.3} (Z.ai, FP8).

\begin{table}[ht]
\centering\footnotesize
\caption{Harness configuration. Entries are the harnesses' own defaults, except the OpenHands context and output limits and the openJiuwen composition described in Section~\ref{sec:harness_config}. Tools are those on TUA-Bench and Terminal-Bench; on ALE-CLI every harness also receives the benchmark's fourteen computer-use tools.}
\label{tab:harness_config}
\setlength{\tabcolsep}{3pt}
\begin{tabular}{p{0.12\linewidth}p{0.11\linewidth}p{0.27\linewidth}p{0.27\linewidth}p{0.13\linewidth}}
\toprule
Harness & Version & Tools & Context management & Turn limit \\
\midrule
OpenHands & SDK 1.44.1 & terminal, file editor, task tracker, think, finish & 1M-token window, no condensation & 500 steps; stuck detector \\
DSH & 0.1.1-rc.2 & 26 standard tools & native compaction at 80\% of the window & none \\
PI & 0.84.4 & read, bash, edit, write & native compaction near the window limit & none \\
openJiuwen & 0.1.18 & read, write, edit, glob, list, grep, bash & native context compression & 8 outer rounds \\
Codex & 0.150.1 & native code-execution tool & native compaction (272K window) & none \\
Claude Code & 2.1.251 & 21 native tools & native auto-compaction (1M window) & none \\
\bottomrule
\end{tabular}
\end{table}

\paragraph{Task collections.}
Terminal-Bench uses the 63-task non-H100 (CPU-only) subset of v4.0.0 (commit \texttt{452bf305c6da}), and ALE-CLI uses 99 local-Docker tasks (commit \texttt{0b6465b13c85}).

\paragraph{Reruns.}
Tasks affected by infrastructure failures, such as environment start-up or Docker readiness failures and provider errors, were rerun. Each task counts its last run; a task without a verifier result in that run counts as zero.

\paragraph{openJiuwen--DeepSeek on ALE-CLI.}
DeepSeek V4 Pro does not accept images, and the provider's API serves image requests with DeepSeek V4.1 Flash. This applied to 343 calls in four image and computer-use tasks.

\section{Code and data release}
\label{app:release}
The code is available at \url{https://github.com/liyix/finding-the-right-fit}. It contains the adapters that run each of the six harnesses on the three benchmarks, the configuration used in this study (harness versions, OpenRouter routes pinned to first-party providers, reasoning effort, and time limits), and the scripts that select the final run of each task, reconcile costs with OpenRouter's per-generation records, and produce the tables and figures in this paper.

The data are available at \url{https://huggingface.co/datasets/yixuanli97/finding-the-right-fit}. They contain a per-task results table and one trajectory for each of the 6,204 scored runs, converted from each harness's native logs into a common chat format with reward, cost, token usage, and agent time. Of these runs, 6,067 include the full model--tool conversation; the remaining 137, mostly runs whose environment never started, carry results only. Secrets and host paths are removed, very long outputs are truncated, and no verifier files or reference data are included. The data are released for analysis of agent behavior and harness comparison under CC BY-NC 4.0; following the benchmarks' own terms, they should not be used to train, fine-tune, or distill models.

\section{Additional resource-use figures}
\label{app:resources}

\begin{figure}[p]
\centering
\includegraphics[width=\linewidth]{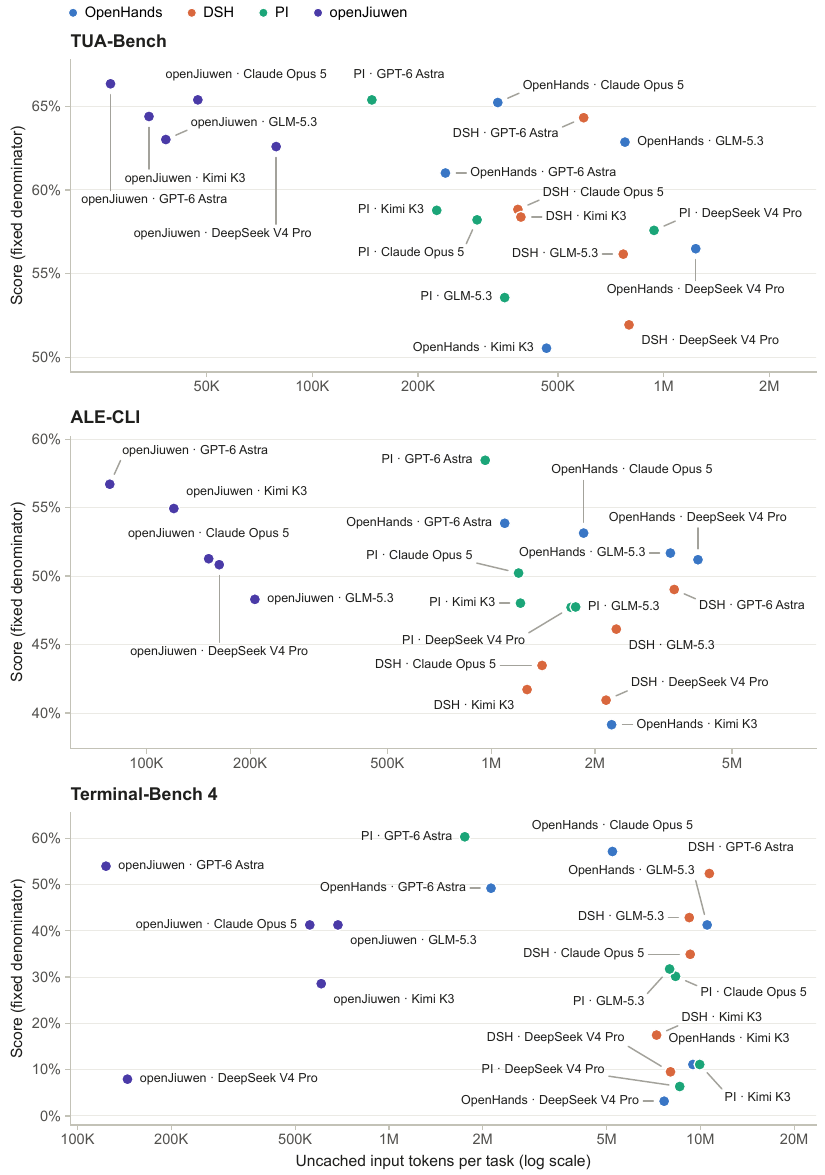}
\caption{Score versus uncached input tokens per task (input minus cache reads, log scale) for the four configurable harnesses.}
\label{fig:prompt_tokens}
\end{figure}

\begin{figure}[p]
\centering
\includegraphics[width=\linewidth]{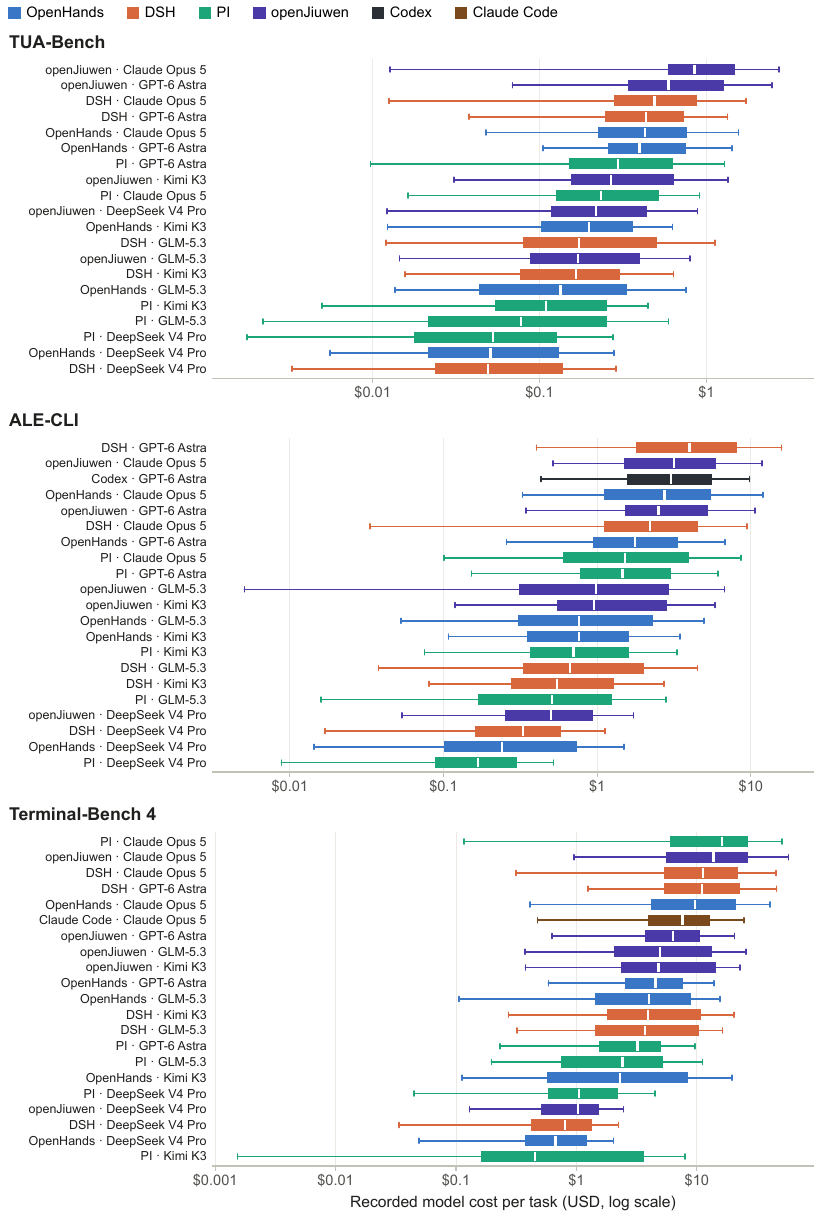}
\caption{Per-task cost distribution by configuration (log scale). Boxes show the median and interquartile range; whiskers extend to 1.5$\times$IQR; outliers are not drawn.}
\label{fig:task_cost_dist}
\end{figure}

\begin{figure}[p]
\centering
\includegraphics[width=\linewidth]{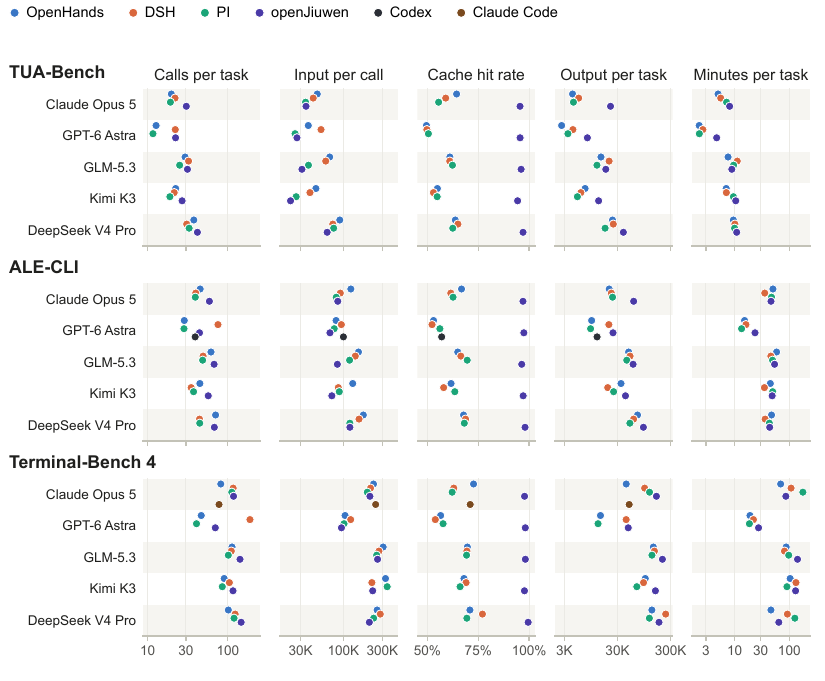}
\caption{Resource use by configuration, per task: model calls, input tokens per call, cache hit rate (cached share of input tokens), output tokens, and agent minutes. Each dot is one harness, and input includes cached tokens. Codex and Claude Code appear where their logs record tokens.}
\label{fig:resource_profile}
\end{figure}

\begin{figure}[p]
\centering
\includegraphics[width=\linewidth]{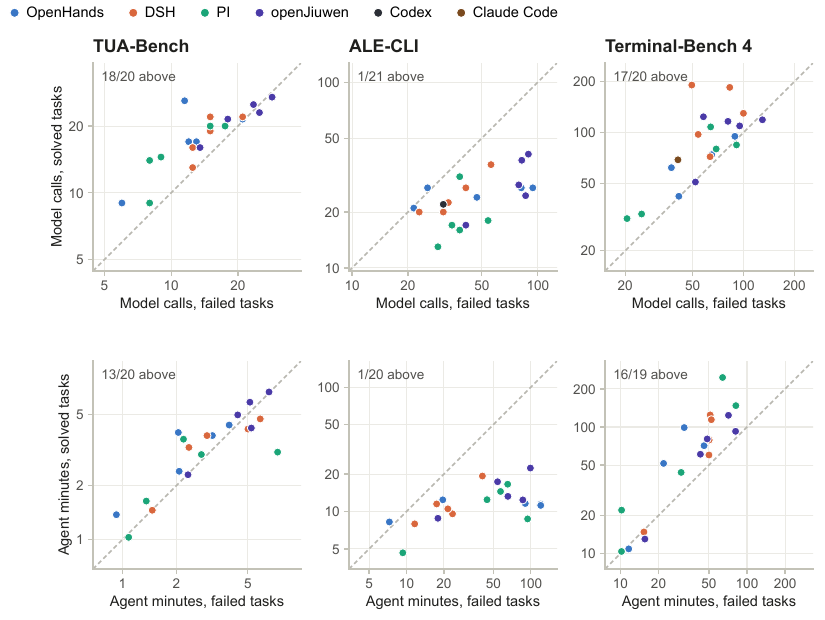}
\caption{Median model calls (top) and agent minutes (bottom) on solved (reward 1) versus failed (reward 0) tasks. Each dot is one configuration with at least three tasks in each group; points above the diagonal spend more on the tasks they solve.}
\label{fig:turns_outcome}
\end{figure}

\section{Cost by configuration}
\label{app:costs}
\begin{table}[ht]
\centering\footnotesize
\caption{Agent-model cost (USD, OpenRouter pricing) of each configuration, summed over the last run of every task. Dividing by the task count (120/99/63) gives the cost per task in Figure~\ref{fig:cost_performance}. OJW denotes openJiuwen; Native is Claude Code for Claude Opus~5 and Codex for GPT-6 Astra.}
\label{tab:cost}
\setlength{\tabcolsep}{4pt}
\begin{tabular}{lrrrrr}
\toprule
Model & OpenHands & DSH & PI & OJW & Native \\
\midrule
\multicolumn{6}{l}{\textit{TUA-Bench} (120 tasks)} \\
Claude Opus 5 & 86.60 & 91.74 & 70.28 & 162.44 & 113.73 \\
GPT-6 Astra & 86.03 & 150.97 & 76.92 & 157.62 & 113.08 \\
GLM-5.3 & 51.86 & 56.47 & 29.82 & 45.15 & -- \\
Kimi K3 & 46.85 & 47.83 & 33.51 & 56.67 & -- \\
DeepSeek V4 Pro & 14.52 & 12.81 & 12.07 & 43.53 & -- \\
\midrule
\multicolumn{6}{l}{\textit{ALE-CLI} (99 tasks)} \\
Claude Opus 5 & 498.26 & 352.20 & 323.34 & 484.45 & 570.21 \\
GPT-6 Astra & 275.51 & 672.49 & 266.97 & 513.90 & 481.88 \\
GLM-5.3 & 204.91 & 160.74 & 138.12 & 195.41 & -- \\
Kimi K3 & 217.04 & 105.47 & 151.43 & 220.40 & -- \\
DeepSeek V4 Pro & 49.20 & 42.06 & 25.73 & 66.83 & -- \\
\midrule
\multicolumn{6}{l}{\textit{Terminal-Bench~4} (63 tasks)} \\
Claude Opus 5 & 814.82 & 1122.42 & 1277.48 & 1208.69 & 780.15 \\
GPT-6 Astra & 340.30 & 1256.48 & 293.30 & 697.58 & 290.35 \\
GLM-5.3 & 465.53 & 411.55 & 376.45 & 717.10 & -- \\
Kimi K3 & 551.35 & 494.85 & 486.79 & 750.68 & -- \\
DeepSeek V4 Pro & 50.59 & 82.43 & 92.32 & 75.69 & -- \\
\bottomrule
\end{tabular}
\end{table}

\end{document}